\documentclass[10pt,twocolumn,letterpaper]{article}

\usepackage{cvpr}              
\definecolor{cvprblue}{rgb}{0.21,0.49,0.74}
\usepackage[pagebackref,breaklinks,colorlinks,allcolors=cvprblue]{hyperref}
\usepackage{algorithmic}
\usepackage{graphicx}
\usepackage{textcomp}
\usepackage{xcolor}
\usepackage{multirow}
\usepackage{booktabs}
\usepackage[table]{xcolor}
\def\paperID{10} 
\def\confName{CVPR}
\def\confYear{2026}

\title{LiteMVS: Efficient Multi-View Stereo with Foundation Distillation and Expert Aggregation}

\author{
Tianbao Zhang\textsuperscript{1,3} \quad
Zeyu Liu\textsuperscript{1} \quad
Shuyu Wu\textsuperscript{1} \quad
Fanxing Li\textsuperscript{1} \\
Zhaoxin Fan\textsuperscript{2,\textdagger} \quad
Wenjun Wu\textsuperscript{2} \quad
Danping Zou\textsuperscript{1,\textdagger} \\
\textsuperscript{1}Shanghai Key Laboratory of Intelligent Sensing and Recognition, Shanghai Jiao Tong University \\
\textsuperscript{2}Beijing Advanced Innovation Center for Future Blockchain and Privacy Computing,\\ School of Artificial Intelligence, Beihang University\\
\textsuperscript{3}Dim12 AI Inc \\
}

\begin{document}
\twocolumn[{
\renewcommand\twocolumn[1][]{#1}
\maketitle

\begin{center}
    \captionsetup{type=figure}
    \includegraphics[width=1.0\textwidth]{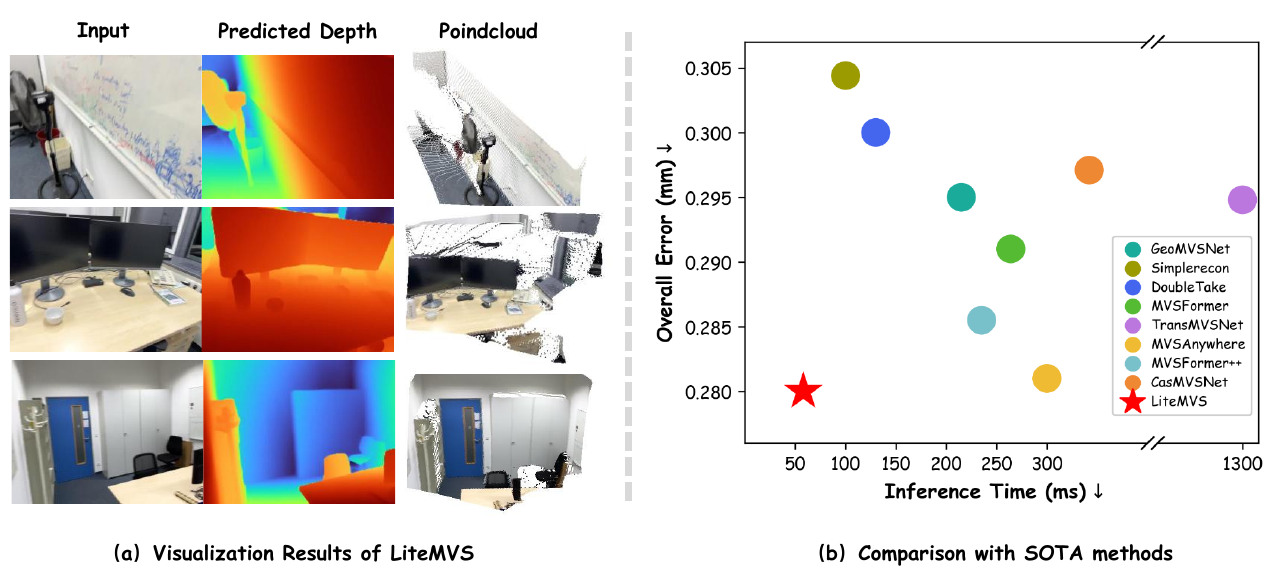}
    \captionof{figure}{(a) Qualitative visualization results of LiteMVS, including the input images, predicted depth maps, and reconstructed point clouds on representative indoor scenes. LiteMVS produces geometrically consistent depth predictions and high-quality point cloud reconstructions with clear scene structures. (b) Comparison with SOTA methods in terms of overall error and inference time, where lower is better in both metrics. LiteMVS achieves the lowest overall error while maintaining the fastest inference speed, demonstrating a favorable trade-off between accuracy and efficiency.}
\vspace{0.5cm}\label{teaser}
\end{center}}]

\begingroup
\renewcommand\thefootnote{$\dagger$}
\footnotetext{Corresponding author.}
\endgroup

\begin{abstract}
Real-time 3D perception is crucial for robotics, augmented reality, and embodied intelligence applications. Existing multi-view stereo (MVS) methods primarily rely on geometric correspondences, which often fail in textureless or repetitive regions, while monocular depth models leverage strong image-level priors but lack robust multi-view geometric constraints. More importantly, in robotics and embodied manipulation scenarios, high-quality 3D geometry is not only essential for static reconstruction, but also serves as a critical foundation for learning temporally consistent 4D representations. To obtain visual representations with stronger structural awareness and greater potential for spatiotemporal extension, we present LiteMVS, a lightweight multi-view depth estimation model that integrates plane-sweep geometric reasoning with strong monocular semantic and structural priors. The central idea of LiteMVS is to efficiently inject high-level monocular knowledge, obtained from lightweight segmentation models and large-scale vision foundation models, into a multi-view stereo framework. In particular, LiteMVS enriches the cost volume with semantic descriptors and employs a Mixture-of-Experts (MoE) formulation to enable adaptive geometric aggregation across depth hypotheses. Moreover, geometric priors distilled from vision foundation models further strengthen monocular guidance without increasing inference cost. Through this design, LiteMVS not only improves depth estimation and 3D reconstruction quality in static scenes, but also provides a more reliable geometric foundation for subsequent temporal modeling and 4D representation learning. Experiments on ScanNetv2 and 7-Scenes demonstrate that LiteMVS achieves high-quality depth prediction and 3D reconstruction while maintaining competitive efficiency. We further validate the effectiveness of its lightweight representations on downstream grasping tasks based on SpatialForcing, showing that the learned geometry-aware representations can also benefit robotic manipulation.

\end{abstract}    
\section{Introduction}
\label{sec:intro}
Real-time depth estimation is fundamental to a wide range of applications, from augmented reality to robotic control~\cite{1,2,fan1,fan2,fan3,fan4}. Among existing solutions, multi-view stereo (MVS) has emerged as a practical paradigm for estimating dense depth from multiple calibrated images at interactive speeds. In recent years, learning-based MVS methods~\cite{3,4} have attracted growing attention by leveraging deep neural networks to improve robustness and accuracy under challenging imaging conditions. These methods typically construct a cost volume by computing pairwise feature correlations between a reference image and multiple source images through differentiable warping, and then infer depth from the aggregated matching evidence. Despite recent advances~\cite{5,mvsformer++} in enhancing cost volume quality through stronger feature extraction, the construction process still inevitably introduces noisy background interference and unreliable matching costs. Prior studies~\cite{simple,doubletake} have shown that lightweight architectures based on 2D CNNs with cost-volume reasoning can achieve competitive depth accuracy while substantially reducing the computational overhead of 3D convolutional approaches. However, existing efficient MVS methods still face a persistent challenge: balancing geometric accuracy with computational efficiency.

At the same time, recent progress in vision foundation models opens a promising direction for overcoming the limitations of efficient MVS. Foundation models for monocular depth estimation and semantic segmentation~\cite{stablenormal,mobilesamv2}, such as Depth Anything v2~\cite{da2} and Segment Anything~\cite{segment}, are trained on Internet-scale data and exhibit remarkable generalization across diverse scenes. Beyond their task-specific outputs, these models encode rich monocular priors, including relative depth ordering, structural regularities, and semantic boundary information. Such cues are highly complementary to the geometric correspondences exploited by MVS: while MVS relies on local appearance matching that often becomes unreliable in textureless or repetitive regions, monocular priors provide global scene understanding and boundary awareness that can help resolve these ambiguities. Moreover, for robotics and embodied systems, robust depth estimation is valuable not only for static 3D reconstruction, but also as a foundation for learning temporally coherent scene representations. In particular, real-world embodied applications increasingly demand real-time 4D representation models that can capture both spatial structure and temporal dynamics for downstream perception and interaction. This raises an important question: how can we design efficient MVS frameworks that incorporate stronger semantic and geometric priors, while preserving the low computational cost required for real-time applications? This question motivates us to revisit efficient MVS from a new perspective. Rather than directly integrating heavy foundation models into the inference pipeline, we seek a principled way to distill their monocular knowledge into a lightweight multi-view architecture.

In this paper, we introduce LiteMVS, a lightweight multi-view depth estimation model that combines plane-sweep geometric reasoning with strong monocular semantic and structural priors. Specifically, LiteMVS injects high-level features from a lightweight segmentation encoder into the cost volume to improve geometric reasoning, especially around object boundaries and textureless regions. While training remains centered on depth estimation, segmentation cues are used as auxiliary semantic guidance and remain optionally available at inference time. To further bridge the gap between compact models and large-scale vision foundation models, we derive pseudo-labels for relative depth and surface normals from foundation model predictions and use them as auxiliary supervision during training. This output-level distillation allows LiteMVS to absorb strong geometric priors without introducing additional inference cost. Furthermore, inspired by Mixture-of-Experts (MoE)~\cite{moe} architectures, we design an MoE-based cost aggregation strategy that enables different experts to specialize in distinct depth regimes while maintaining efficiency. In this way, LiteMVS not only improves the accuracy-efficiency trade-off of lightweight MVS, but also provides a stronger geometric basis for downstream spatiotemporal representation learning and embodied perception.

We evaluate LiteMVS on the ScanNetv2 and 7-Scenes datasets, where it achieves consistent improvements over existing lightweight multi-view stereo methods. The model delivers accurate depth prediction and high-quality 3D reconstruction while maintaining competitive efficiency. These results demonstrate the effectiveness of integrating semantic guidance, adaptive expert aggregation, and knowledge distillation for efficient multi-view depth estimation. As shown in Fig.~\ref{teaser}, we present both qualitative visualization results of LiteMVS, including predicted depth maps and reconstructed point clouds, and a quantitative comparison with state-of-the-art methods in terms of overall error and inference time, showing that LiteMVS achieves promising reconstruction quality while maintaining fast inference speed. Our contributions can be summarized as follows:

\begin{itemize}
    \item We introduce a \textbf{semantic-aware feature volume} that incorporates features from a lightweight segmentation encoder as auxiliary guidance, enabling the model to leverage high-level semantic cues for more robust depth estimation, particularly around object boundaries and textureless regions.
    
    \item We present a \textbf{Mixture-of-Experts (MoE) aggregation module} that adaptively compresses the 4D feature volume into a compact 3D cost representation. This expert-driven mechanism enhances geometric modeling across varying depth ranges while preserving inference efficiency.
    
    \item We conduct a \textbf{knowledge distillation scheme} that transfers geometric priors—including relative depth and surface normals—from large-scale vision foundation models via pseudo-label supervision. This enables LiteMVS to inherit strong generalization capabilities with no additional inference cost.
\end{itemize}

\begin{figure*}[t]
\centering
\includegraphics[width=1.0\linewidth,height=12cm]{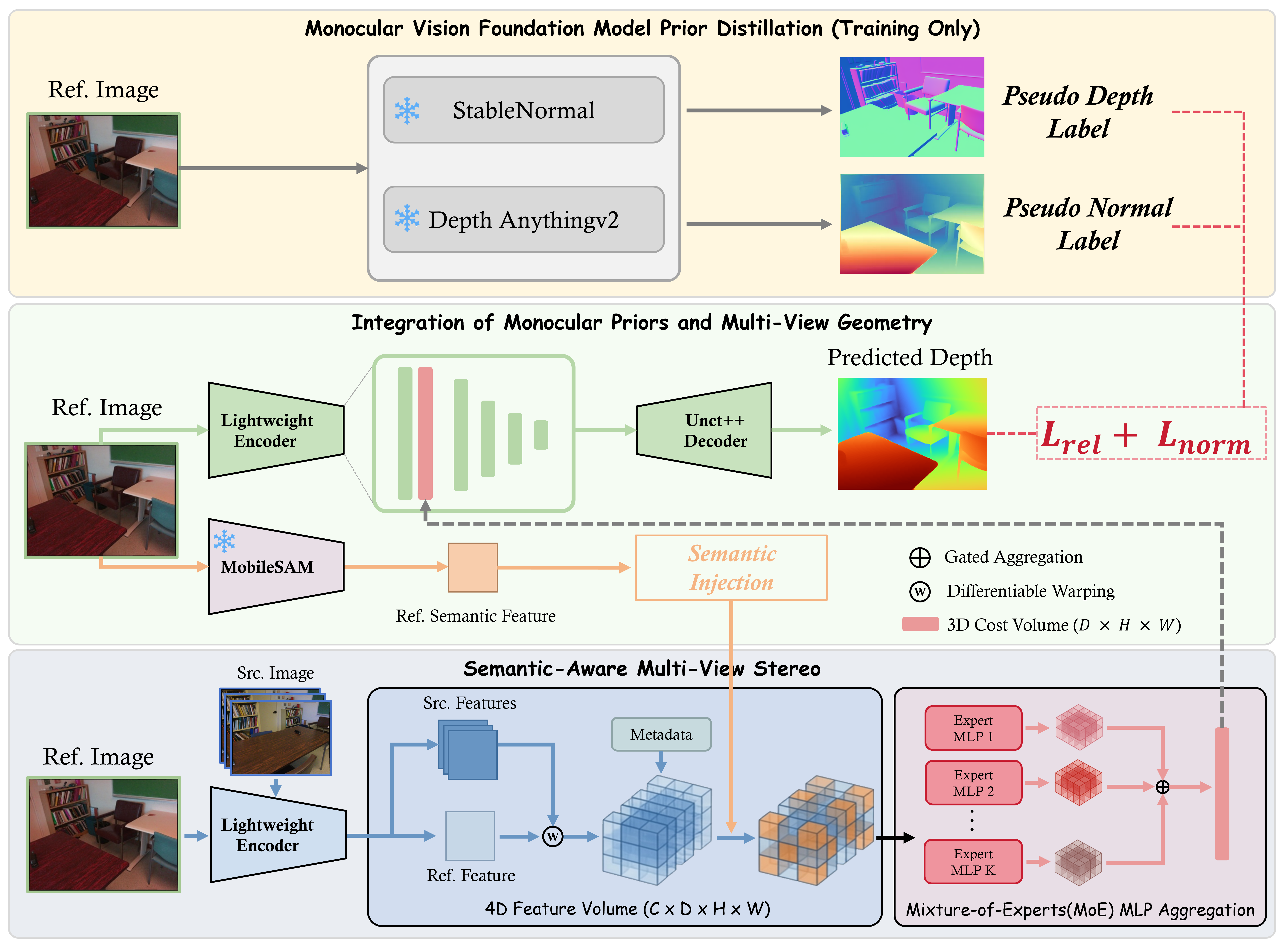}

\caption{\textbf{Framework of LiteMVS.} The pipeline consists of three components: (1) monocular vision foundation model distillation that generates pseudo depth and normal labels for training supervision; (2) semantic feature injection via MobileSAM to enhance the cost volume with scene-level context; and (3) semantic-aware multi-view stereo with a Mixture-of-Experts MLP for adaptive feature aggregation and robust depth prediction.}\label{fig2}
\end{figure*}

\section{Related Work}
\label{sec:related_work}

\subsection{Multi-View Stereo}
Multi-view stereo (MVS) estimates depth from multiple posed images by exploiting epipolar geometry. Recent approaches~\cite{3,4} have increasingly adopted deep neural networks for end-to-end depth inference, typically using plane-sweep cost volumes. Subsequent research~\cite{5,mvsformer++} explored improved network architectures, robustness to occlusions and dynamic scenes, and computational efficiency. Beyond appearance-based matching, several works~\cite{simple,doubletake} have attempted to enrich cost volumes with auxiliary cues such as viewing direction, relative camera pose, or temporal information to improve depth estimation. While these approaches demonstrate that auxiliary metadata can be beneficial, the incorporated cues are typically limited to low-level geometric attributes and are often introduced in a fixed or shallow manner. However, they remain insufficient to resolve ambiguities in challenging regions such as object boundaries or textureless surfaces, where higher-level semantic and structural understanding is essential. In this work, we go beyond geometric metadata by incorporating rich semantic representations from a lightweight segmentation encoder, enabling the model to leverage high-level scene understanding for more robust depth estimation.

\subsection{Single-View Depth Estimation}

Monocular depth estimation methods~\cite{6,7} infer depth from a single image without relying on multi-view observations. Recent advances have focused on developing general-purpose models that generalize across diverse scenes by scaling model capacity and training data~\cite{da2,da}, often combining real and synthetic datasets with stronger image-level priors. However, models trained without explicit metric supervision typically predict relative rather than metric depth, and monocular approaches remain fundamentally constrained by single-view inference even when multi-view information is available. Despite these limitations, modern monocular models capture rich semantic and structural cues that can provide valuable guidance. Motivated by this, we explored combining monocular representations with multi-view cost volumes, leveraging the complementary strengths of monocular priors and multi-view geometric reasoning.
Despite these limitations, modern monocular models capture rich geometric and structural knowledge that can serve as valuable guidance. This motivates us to distill such priors into lightweight MVS frameworks.

\section{Method}
\label{sec:method}

We introduce LiteMVS, a semantic-aware multi-view depth estimation framework that integrates plane-sweep geometric reasoning with strong monocular priors. Our design is motivated by the complementary nature of multi-view geometry and monocular cues: plane-sweep matching enforces multi-view consistency, while monocular priors provide structural understanding often absent in correspondence-driven pipelines. To bridge these paradigms, LiteMVS incorporates: (1) semantic features from a lightweight segmentation encoder injected into the cost volume for boundary-sensitive guidance; (2) a Mixture-of-Experts (MoE) formulation for adaptive aggregation across distinct depth regimes; and (3) geometric priors distilled from vision foundation models via pseudo-label supervision, transferring rich geometric knowledge at no extra inference cost. An overview is provided in Fig.~\ref{fig2}.

\subsection{Problem Setup}
\label{sec:problem_setup}

Given a set of posed RGB images $\mathcal{I}=\{I_0, I_1, \ldots, I_N\}$ with known camera intrinsics $K_i$ and relative extrinsics $(R_{0,i}, t_{0,i})$, where $I_0$ is the reference view, our goal is to predict a dense depth map $\hat{D}_0$ for the reference image. 

\subsection{Semantic-Aware 4D Feature Volume}
\label{sec:volume}

\noindent\textbf{Visual matching features.}
We encode each image with a lightweight CNN backbone $\phi$ to obtain feature maps for cross-view matching: $F_i=\phi(I_i)$. These features form the basis for plane-sweep alignment and multi-view correspondence reasoning.

\noindent\textbf{Semantic descriptors as monocular priors.}
To inject high-level structure and boundary cues, the reference image $I_0$ is additionally encoded by a lightweight segmentation encoder $\psi$ (e.g., MobileSAM), producing semantic descriptors $S_0=\psi(I_0)$. LiteMVS does not train a segmentation head; $S_0$ serves only as an auxiliary prior that complements appearance-based matching, especially around object boundaries and in textureless regions.\\
\noindent\textbf{Plane-sweep warping and volume construction.}
Given a set of depth hypotheses $\{d_k\}_{k=1}^{D}$, we construct a plane-sweep feature volume by warping each source-view feature map into the reference view via differentiable homography:
\begin{equation}
F_{i\rightarrow 0}(d_k)=\mathrm{warp}\!\left(F_i, K_i, R_{0,i}, t_{0,i}, d_k\right).
\end{equation}
The warped features are aggregated across source views to form a 4D feature volume $V\in\mathbb{R}^{C\times D\times H\times W}$. Following~\cite{simple}, we augment $V$ with geometric metadata $M$. To incorporate monocular priors, semantic descriptors $S_0$ are broadcast along the depth dimension and concatenated to form the semantic-aware volume $\tilde{V}=\mathrm{concat}(V, M, S_0)$,
which enables the model to jointly leverage multi-view consistency and monocular structural cues for depth estimation.

\subsection{Mixture-of-Experts Cost Aggregation}
\label{sec:moe}

In ~\cite{simple}, the high-dimensional 4D feature volume is reduced using a single shared MLP applied uniformly across all depth hypotheses. To capture diverse geometric patterns across varying depth ranges, LiteMVS adopts a Mixture-of-Experts (MoE) formulation for adaptive cost aggregation.

Given a voxel feature $\tilde{V}_{:,d,h,w}$, multiple expert MLPs independently transform the input, and a lightweight gating network predicts soft expert weights. The aggregated cost is computed as:
\begin{equation}
C_{d,h,w} = \sum_{j=1}^{K} g_j \cdot f_j(\tilde{V}_{:,d,h,w}),
\end{equation}
where $f_j$ denotes the $j$-th expert and $g_j$ its corresponding weight. This formulation allows different experts to specialize in distinct depth regimes, improving representational flexibility while preserving efficiency.

\subsection{Pseudo-Label Distillation from Vision Foundation Models}
\label{sec:distill}

To strengthen monocular priors without increasing inference cost, we distill geometric knowledge from vision foundation models via pseudo-label supervision. Specifically, we generate relative depth pseudo-labels $\tilde{D}^{*}$ using Depth Anything V2~\cite{da2} and surface normal pseudo-labels $\tilde{N}^{*}$ using StableNormal~\cite{stablenormal}. These pseudo-labels provide rich geometric guidance that complements multi-view supervision.

\noindent\textbf{Relative depth supervision.}
Since foundation models typically predict relative rather than metric depth, we adopt a scale-invariant loss to align predicted depth with pseudo-labels:
\begin{equation}
\mathcal{L}_{\mathrm{rel}}=\mathrm{SI}(\hat{D}_0,\tilde{D}^{*}).
\end{equation}

\noindent\textbf{Surface normal supervision.}
We convert the predicted depth map into surface normals via a differentiable depth-to-normal operator $\mathcal{N}(\cdot)$. The normal loss combines SSIM and gradient terms to enforce both structural similarity and edge consistency with pseudo-labels:
\begin{equation}
\mathcal{L}_{\mathrm{norm}}=\mathcal{L}_{\mathrm{SSIM}}(\hat{N}_0, \tilde{N}^{*}) + \lambda_{\mathrm{grad}}\mathcal{L}_{\mathrm{grad}}(\hat{N}_0, \tilde{N}^{*}),
\end{equation}
where $\hat{N}_0 = \mathcal{N}(\hat{D}_0)$ denotes the predicted surface normals.

By combining relative depth and surface normal supervision, the model learns to capture both global geometric structure and local surface details from foundation model priors. As shown in Fig.~\ref{fig:normal_vis}, incorporating pseudo-label distillation leads to noticeably sharper depth predictions, particularly around object boundaries and fine-grained structures.

\subsection{Loss Function}
\label{sec:loss}

Our training objective builds upon the loss formulation of SimpleRecon~\cite{simple}, which serves as our primary supervision for multi-view depth estimation. On top of this baseline objective, we further introduce the proposed distillation losses to transfer geometric priors from vision foundation models. The overall loss is defined as
\begin{equation}
    \mathcal{L}=\mathcal{L}_{\text{simple}}+\lambda_{\mathrm{rel}}\mathcal{L}_{\mathrm{rel}}+\lambda_{\mathrm{norm}}\mathcal{L}_{\mathrm{norm}},
\end{equation}
where $\lambda_{\mathrm{rel}}$ and $\lambda_{\mathrm{norm}}$ are balancing weights, and $\mathcal{L}_{\text{simple}}$ denotes the original multi-view supervision objective adopted from~\cite{simple}. The full formulation of $\mathcal{L}_{\text{simple}}$ is provided in the supplementary material.

\section{Experiments}
\label{sec:exp}

\begin{table*}[t]
    \centering
    \caption{\textbf{Depth evaluation.} For each metric, the best-performing method is marked in red, the second-best in orange, and the third-best in yellow. Results for previous methodswere taken from[12], or evaluated for each method using their keyframes.}
    \label{tab:comparison}
    \scalebox{0.75}{
    \begin{tabular}{c|ccccc|ccccc}
        \toprule
        \multirow[c]{2}{*}{\raisebox{-1ex}{Method}}& \multicolumn{5}{c|}{ScanNetv2} & \multicolumn{5}{c}{7Scenes} \\
        \cmidrule(lr){2-6} \cmidrule(lr){7-11}
        & Abs Diff$\downarrow$ & Abs Rel$\downarrow$ & Sq Rel$\downarrow$ & $\delta < 1.05$ $\uparrow$ & $\delta < 1.25$ $\uparrow$ 
        & Abs Diff$\downarrow$ & Abs Rel$\downarrow$ & Sq Rel$\downarrow$ & $\delta < 1.05$ $\uparrow$ & $\delta < 1.25$ $\uparrow$ \\
        \midrule
        DPSNet~\cite{dpsnet} & 0.1552 & 0.0795 & 0.0299 & 49.36 & 93.27 & 0.1966 & 0.1147 & 0.0550 & 38.81 & 87.07 \\
        MVDepthNet~\cite{mvdepthnet} & 0.1648 & 0.0848 & 0.0343 & 46.71 & 92.77 & 0.2009 & 0.1161 & 0.0623 & 38.81 & 87.70 \\
        GPMVS~\cite{multi} & 0.1494 & 0.0757 & 0.0252 & 51.04 & 93.96 & 0.1739 & 0.1003 & 0.0462 & 42.71 & 90.32 \\
        DeepVideoMVS, fusion~\cite{deepvideomvs} & 0.1186 & 0.0583 & 0.0190 & 60.20 & 96.76 & 0.1448 & 0.0828 & 0.0335 & 47.96 & 93.79 \\
        \rowcolor{yellow!30}Simplerecon~\cite{simple} & 0.0873 & 0.0430 & 0.0128 & 74.12 & 98.05 & 0.1045 & 0.0575 & 0.0156 & 60.12 & \cellcolor{orange!30}97.33\\
        \rowcolor{orange!30}DoubleTake~\cite{doubletake} & 0.0767 & 0.0369 & 0.0112 & 79.94 & 98.35 & 0.0985 & 0.0534 & 0.0156 & 64.76 & \cellcolor{yellow!30}97.01 \\
        \midrule
        \rowcolor{red!30}\textbf{LiteMVS (Ours)} &  \textbf{0.0702} & \textbf{0.0311} & \textbf{0.0097} & \textbf{82.45} & \textbf{98.52} & \textbf{0.0898} & \textbf{0.0480} & \textbf{0.0143} & \textbf{69.47} & \textbf{97.52} \\
        \bottomrule
    \end{tabular}%
    }
\end{table*}

\begin{figure*}[h]
\centering
\includegraphics[width=1.0\linewidth,height=10cm]{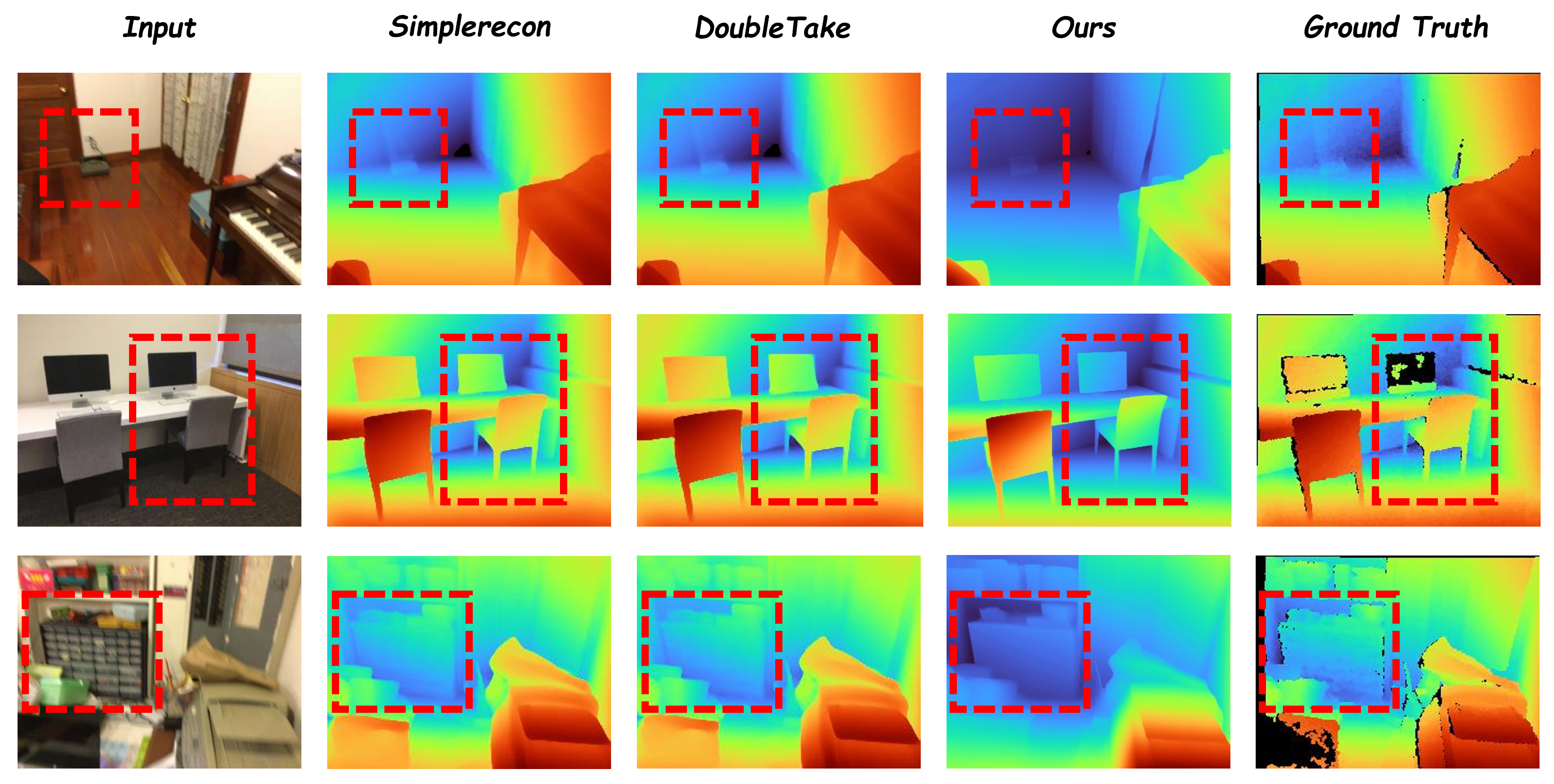}
\caption{\textbf{Qualitative results of LiteMVS.} LiteMVS produces accurate depth predictions with sharp object boundaries and consistent surface geometry, enabling high-quality 3D reconstruction across diverse indoor scenes. 
Compared to existing methods, fine-grained geometric details and boundary regions are better preserved, as highlighted in the red boxes.}\label{fig:depth_vis}
\end{figure*}

We train our models on a combination of ScanNetv2~\cite{scannet} and ScanNet++~\cite{scannet++} datasets, which together provide large-scale indoor RGB-D sequences captured with handheld sensors. We evaluate our method on the ScanNetv2 test set and 7-Scenes~\cite{scene}. For 7-Scenes, we directly apply the trained models without fine-tuning, using the test split proposed in~\cite{deepvideomvs} to evaluate cross-dataset generalization. Additional implementation details provided in the supplementary material.

\subsection{Depth Estimation}
We present quantitative depth estimation results in Table~\ref{tab:comparison}, evaluated using the standard metrics proposed by Eigen et al.~\cite{eigen}. 
In addition to commonly reported thresholds, we further adopt a stricter accuracy criterion with $\delta < 1.05$ to better differentiate strong-performing models. 
Our approach is directly compared with previously works, including DeepVideoMVS~\cite{deepvideomvs}, on both ScanNetv2 and 7-Scenes.
For evaluation, we use the official test split for ScanNetv2 and follow the test split introduced in~\cite{deepvideomvs} for 7-Scenes. 
Depth metrics are computed independently for each keyframe and averaged across all keyframes within each test set.

\begin{figure}[h]
\centering
\includegraphics[width=1.0\linewidth,height=4cm]{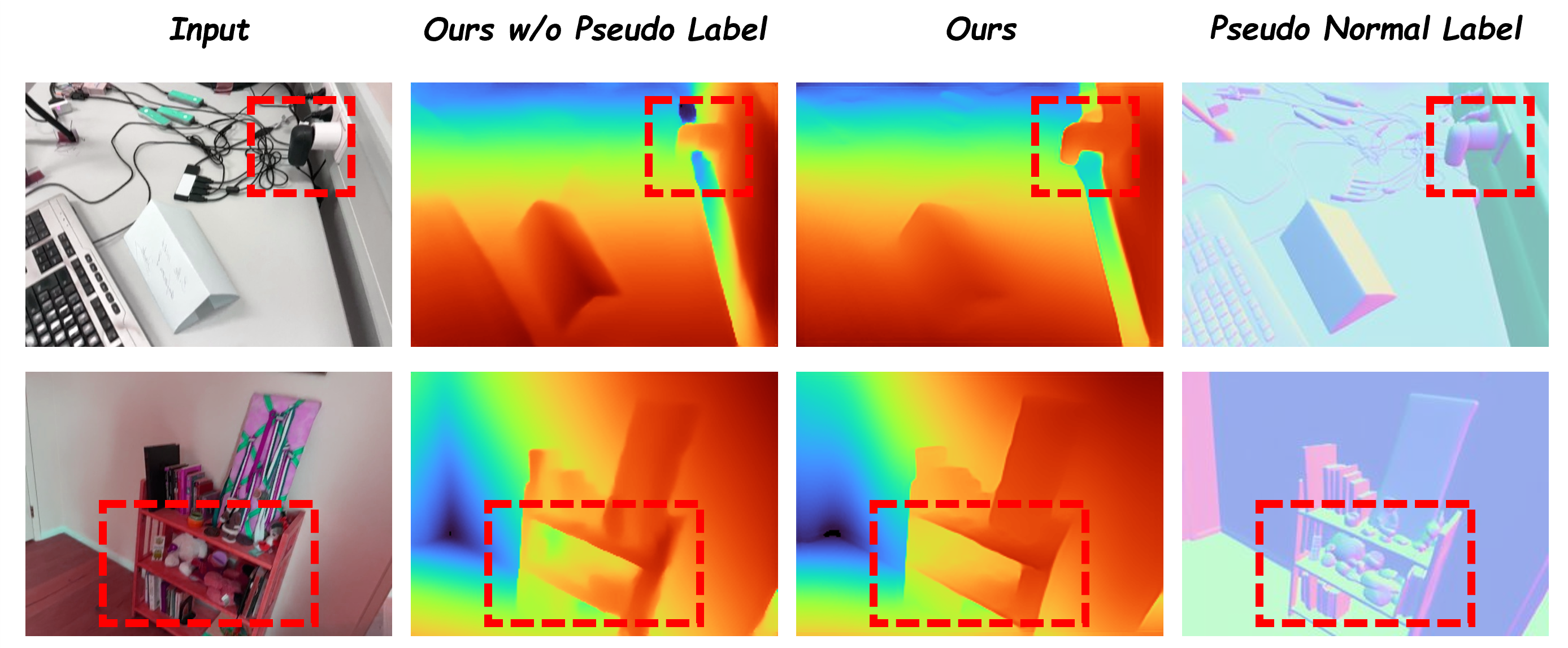}
\caption{\textbf{Effectiveness of Pseudo-Label Distillation.}
Compared with the model trained without pseudo-label supervision, LiteMVS produces sharper object boundaries and more coherent depth transitions (highlighted by red boxes).
}\label{fig:normal_vis}
\end{figure}

Our model outperforms existing baselines across all evaluated depth estimation metrics. By injecting high-level semantic descriptors and adaptive cost aggregation mechanisms into the cost volume, LiteMVS effectively resolves ambiguities in challenging regions such as textureless surfaces and object boundaries. Qualitative results of the predicted depth maps are shown in Fig.~\ref{fig:depth_vis}, validating the improvements of our method in terms of geometric accuracy and structural consistency. Our method yields more accurate geometry and improved structural consistency compared to prior approaches, with noticeably cleaner boundaries and fewer artifacts in ambiguous regions. Fig.~\ref{fig:normal_vis} further illustrates that, with the distillation of geometric priors from powerful vision foundation models, LiteMVS yields more refined and detailed depth predictions.

\subsection{3D Reconstruction Quality and Efficiency}

We assess both reconstruction quality and efficiency under a unified evaluation protocol. 
Reconstruction accuracy is measured using the standard procedure introduced in TransformerFusion~\cite{transformerfusion}, which employs a ground-truth mesh-based mask to avoid penalizing valid geometry not present in the reference mesh. 
Quantitative results are summarized in Table~\ref{tab:recon_latency}. 
Although LiteMVS employs a lightweight depth-based reconstruction pipeline without global volumetric refinement, it achieves state-of-the-art performance across multiple reconstruction metrics, demonstrating superior geometric fidelity with substantially reduced model complexity.

Beyond reconstruction quality, we evaluate computational efficiency in terms of per-frame update latency, which is critical for online and interactive applications. Following prior work~\cite{deepvideomvs}, we report the per-frame integration time given a new RGB observation, excluding any waiting time required to satisfy keyframe constraints for fair comparison. As shown in the Table~\ref{tab:recon_latency}, LiteMVS achieves comparable latency to SimpleRecon and DoubleTake while delivering notably better reconstruction quality, striking an effective balance between efficiency and accuracy.

\section{Ablation Study}
We perform ablation studies on ScanNetv2 to evaluate the contribution of the main components in LiteMVS, with results summarized in Table~\ref{tab:ablation}.

\subsection{Effect of Semantic Priors}

Removing semantic descriptors from the 4D feature volume leads to a clear degradation in performance, as shown in Table~\ref{tab:ablation}. 
In particular, incorporating semantic priors reduces the absolute relative error by approximately 25\% and the absolute difference error by over 15\%. 
These improvements indicate that semantic information provides effective structural guidance for depth estimation.

\begin{table}[h]
    \centering
    \caption{Mesh Evaluation and Frame integration latencies for 3D reconstruction. }
    \label{tab:recon_latency}
    \scalebox{0.65}{
    \begin{tabular}{lccccc}
        \toprule
        Method & Comp$\downarrow$ & Acc$\downarrow$ & Recall$\uparrow$ & F-Score$\uparrow$ &  Update Latency (ms)$\downarrow$ \\
        \midrule
        NeuralRecon~\cite{neuralrecon} & \cellcolor{orange!30}5.09 & 9.13 & 0.612 & 0.619 & 90\\
        TransformerFusion~\cite{transformerfusion} & 5.52 & 8.27 & 0.600 & 0.655 & 326\\
        VoRTX~\cite{vortx} & \cellcolor{red!30}4.31 & 7.23 & 0.651 & \cellcolor{yellow!30}0.703 & 4550\\
        Simplerecon~\cite{simple} & 5.53 & \cellcolor{yellow!30}6.09 & \cellcolor{yellow!30}0.658 & 0.671 & \cellcolor{yellow!30}72\\
        DoubleTake~\cite{doubletake} & 5.49 & \cellcolor{orange!30}4.70 & \cellcolor{orange!30}0.701 & \cellcolor{orange!30}0.714 & \cellcolor{red!30}76\\
        \midrule
        \textbf{Ours} & \cellcolor{orange!30}\textbf{5.37} & \cellcolor{red!30}\textbf{4.16} & \cellcolor{red!30}\textbf{0.715} & \cellcolor{red!30}\textbf{0.715} & \cellcolor{orange!30}\textbf{75}\\
        \bottomrule
    \end{tabular}%
    }
\end{table}


\begin{table}[h]
    \centering
    \caption{\textbf{Ablation Evaluation.} Ablation evaluation on depth metrics using DVMVS~\cite{deepvideomvs} keyframes for ScanNet. The best-performing method is marked in red, the second-best in orange, and the third-best in yellow.}
    \label{tab:ablation}
    \scalebox{0.75}{
    \begin{tabular}{c|cccc}
        \toprule
        \multirow[c]{2}{*}{\raisebox{-1ex}{Method}}& \multicolumn{4}{c}{ScanNetv2} \\
        \cmidrule(lr){2-5}
        & Abs Diff$\downarrow$ & Abs Rel$\downarrow$ & Sq Rel$\downarrow$ & $\delta < 1.25$ $\uparrow$ \\
        \midrule
        Ours w/o Sematic & 0.0830 & 0.0412 & 0.0121 & 98.12 \\
        \rowcolor{yellow!30}Ours w/o MoE MLP & 0.0741 & 0.0357 & 0.0110 & 98.29\\
        \rowcolor{orange!30}Ours w/o $L_{MonoKD}$ & 0.0726 & 0.0338 & 0.0104 & 98.34\\
        \midrule
        \rowcolor{red!30}\textbf{LiteMVS(Ours)} &  \textbf{0.0702} & \textbf{0.0311} & \textbf{0.0097} & \textbf{98.52} \\
        \bottomrule
    \end{tabular}%
    }
\end{table}

\begin{table}[h]
    \centering
    \caption{\textbf{Effect of the Number of MoE Experts.} Ablation results of using different numbers of experts in the proposed MoE-based cost aggregation module on ScanNetv2. The best, second-best, and third-best results are highlighted in red, orange, and yellow, respectively.}
    \label{tab:ablation_expert}
    \scalebox{0.7}{
    \begin{tabular}{c|cccc}
        \toprule
        \multirow[c]{2}{*}{\raisebox{-1ex}{Number of MoE Experts}}& \multicolumn{4}{c}{ScanNetv2} \\
        \cmidrule(lr){2-5}
        & Abs Diff$\downarrow$ & Abs Rel$\downarrow$ & Sq Rel$\downarrow$ & $\delta < 1.25$ $\uparrow$ \\
        \midrule
        8  & 0.0789 & 0.0386 & 0.0117 & 98.21 \\
        \rowcolor{yellow!30}5  & 0.0763 & 0.0369 & 0.0112 & 98.27 \\
        \rowcolor{orange!30} 2  & 0.0738 & 0.0345 & 0.0106 & 98.36 \\
        \midrule
        \rowcolor{red!30}\textbf{3 (Ours)} &  \textbf{0.0702} & \textbf{0.0311} & \textbf{0.0097} & \textbf{98.52} \\
        \bottomrule
    \end{tabular}%
    }
\end{table}

\begin{table*}[t]
\centering
\caption{\textbf{Target representation comparison on the LIBERO benchmark.}}
\label{tab:target_representation}
\scalebox{0.9}{
\begin{tabular}{l|c|ccccc}
\toprule
Target Representation & Inference Speed(ms) & Spatial SR (\%) & Object SR (\%) & Goal SR (\%) & Long SR (\%) & Average SR (\%) \\
\midrule
SigLIP\cite{siglip}       & 155 & 95.2 & 94.8  & 94.0 & 91.8 & 94.0 \\
DINOv2~\cite{dinov2}      & 186 & 93.4 & 95.2  & 93.8 & 93.8 & 94.1 \\
VGGT w/o PE  & 201 & \textbf{97.8} & \textbf{100.0} & 96.6 & 84.4 & 94.7 \\
VGGT~\cite{vggt}         & 201 & 97.2 & 99.2  & \textbf{96.8} & \textbf{94.2} & \textbf{96.9} \\
\midrule
LiteMVS (Our)  & \textbf{74} & 95.0 & 94.5  & 93.5 & 92.1 & 94.1 \\
\bottomrule
\end{tabular}}
\end{table*}

\begin{figure*}[h]
\centering
\includegraphics[width=0.9\linewidth,height=5.8cm]{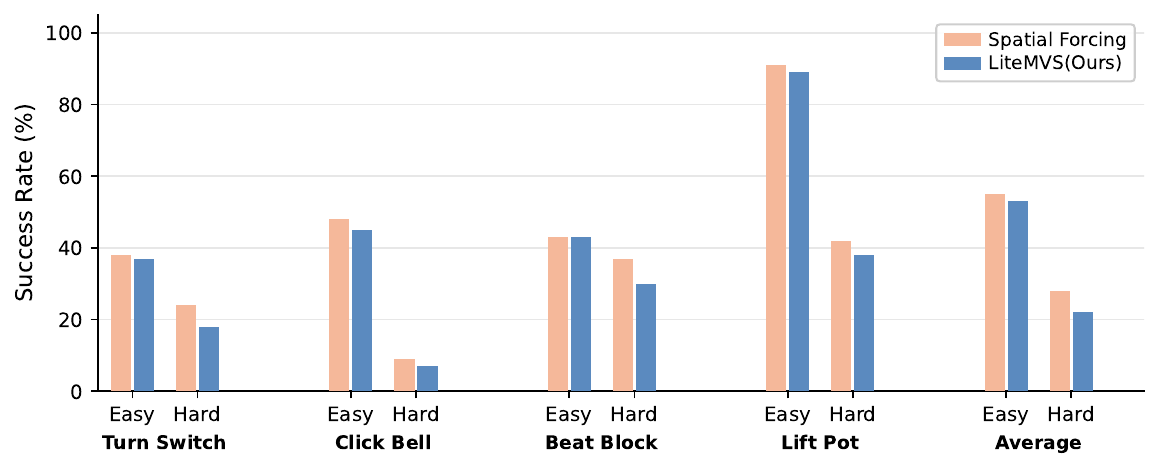}
\caption{\textbf{Comparisons with SpatialForcing~\cite{sf} on RoboTwin 2.0 benchmark.}}\label{fig:robotwin}
\end{figure*}

\subsection{Effect of Mixture-of-Experts Aggregation}

Replacing the Mixture-of-Experts (MoE) MLP with a single shared MLP leads to consistent performance degradation, as shown in Table~\ref{tab:ablation}. 
In particular, introducing the MoE aggregation reduces the absolute relative error by approximately 13\% compared to the single-MLP variant, along with noticeable improvements in squared relative error. 
These gains indicate that the MoE formulation enables more effective aggregation across different depth regimes and geometric contexts. 

\subsection{Effect of Vision Foundation Model Distillation}

We further ablate the proposed MoE-based cost aggregation module by varying the number of experts. Table~\ref{tab:ablation_expert} shows that a moderate number of experts achieves the best performance, while too many experts bring limited gains. Additional studies on expert homogeneity and parameter size are deferred to the supplementary material.

\subsection{Analysis of MoE Design Choices}

We further ablate the proposed MoE-based cost aggregation module by varying the number of experts. As shown in Table~\ref{tab:ablation_expert}, a moderate number of experts gives the best performance, while introducing too many experts brings only limited gains. Additional ablations on expert parameter size are provided in the supplementary material, which further support that the improvement mainly comes from more effective expert specialization and adaptive aggregation, rather than simply increasing model capacity.
\section{Applications on Embodied Task}
\label{sec:application}

To evaluate the transferability of the learned representations, we further apply LiteMVS to downstream embodied manipulation. Specifically, we adopt SpatialForcing\cite{sf} as the evaluation framework and use LiteMVS features as the target visual representation for alignment. We build on OpenVLA-OFT~\cite{openvla} as the base model and conduct experiments on the LIBERO~\cite{libero} benchmark under limited computational resources using a single H100 GPU.

We focus on target representation comparison to examine whether downstream performance mainly depends on the alignment paradigm itself or on the quality of the target representation. To this end, we compare LiteMVS with several widely used visual backbones, including SigLIP~\cite{siglip}, DINOv2~\cite{dinov2}, and VGGT~\cite{vggt}. SigLIP provides strong semantic understanding through image-text alignment, while DINOv2 offers improved visual grounding with fine-grained spatial representations. VGGT, trained on 2D--3D paired data, exhibits strong 3D-aware perception ability. In contrast, LiteMVS provides a lightweight yet geometry-aware representation learned explicitly from efficient multi-view depth estimation.

As shown in Table~\ref{tab:target_representation}, stronger target representations consistently improve manipulation performance over purely 2D visual backbones, confirming that visual embedding alignment is an effective way to enhance spatial perception in embodied tasks. Although LiteMVS is substantially more lightweight than VGGT-based representations, it achieves competitive performance on LIBERO, approaching that of the strongest 3D-aware target representations. Notably, LiteMVS delivers the fastest inference speed among all compared methods, underscoring its clear efficiency advantage for real-time applications. These results suggest that the geometric structure captured by LiteMVS generalizes beyond reconstruction and serves as an effective visual prior for embodied decision-making.

We further compare LiteMVS with SpatialForcing~\cite{sf} on the RoboTwin 2.0 benchmark~\cite{robotwin} in Fig.~\ref{fig:robotwin}. LiteMVS achieves comparable success rates across both easy and hard settings, demonstrating that a lightweight MVS-based representation can support object-centric spatial reasoning and manipulation at a level close to methods that rely on stronger VGGT priors. LiteMVS achieves competitive downstream performance while providing a much more efficient representation and is better aligned with the need for real-time 4D perception in embodied systems, making it a practical, scalable, and highly efficient visual foundation for real-world manipulation task.

\section{Conclusion}
We present LiteMVS, an efficient multi-view stereo framework that enhances lightweight plane-sweep depth estimation through semantic guidance, adaptive Mixture-of-Experts aggregation, and geometric prior distillation from vision foundation models. By integrating semantic descriptors into the cost volume and leveraging pseudo relative depth and normal supervision, LiteMVS effectively combines multi-view geometric reasoning with monocular structural priors without additional inference cost. Experiments on ScanNetv2 and 7-Scenes demonstrate strong depth accuracy, high-quality 3D reconstruction, and a favorable accuracy--efficiency trade-off over prior lightweight MVS methods. Furthermore, the learned geometry-aware representations transfer effectively to embodied manipulation, achieving competitive performance with significantly improved efficiency, highlighting the potential of lightweight multi-view models as practical visual foundations for embodied perception and spatiotemporal representation learning.

\noindent\textbf{Limitations.} LiteMVS remains challenged in scenarios with inherently ambiguous geometry, such as reflective surfaces, transparent objects, and low-texture regions. In addition, while the learned 4d representations show promising transferability, their performance on downstream tasks is not yet optimal, leaving room for improvement.
{
    \small
    \bibliographystyle{ieeenat_fullname}

}


\end{document}